\documentclass{article}

\usepackage[final,nonatbib]{neurips_2022}

\usepackage[utf8]{inputenc} 
\usepackage[T1]{fontenc}    
\usepackage{hyperref}       
\usepackage{url}            
\usepackage{booktabs}       
\usepackage{amsfonts}       
\usepackage{nicefrac}       
\usepackage{microtype}      
\usepackage{xcolor}         
\usepackage{algorithm}
\usepackage{algpseudocode}
\usepackage{multirow}
\usepackage{amsmath}
\usepackage{amssymb}
\usepackage{cleveref}

\DeclareMathOperator*{\argmin}{arg\,min}

\usepackage{pgfplots}
\DeclareUnicodeCharacter{2212}{−}
\usepgfplotslibrary{groupplots,dateplot}
\usetikzlibrary{patterns,shapes.arrows,external}
\pgfplotsset{compat=newest}
\usepackage{tikz}

\pgfplotsset{compat=1.11,
    /pgfplots/ybar legend/.style={
    /pgfplots/legend image code/.code={%
       \draw[##1,/tikz/.cd,yshift=-0.25em]
        (0cm,0cm) rectangle (3pt,0.8em);},
   },
}

\title{Momentum-based Weight Interpolation of Strong Zero-Shot Models for Continual Learning}

\author{%
  Zafir Stojanovski$^{1,}$\thanks{Denotes equal contribution. } ,  Karsten Roth$^{1, \ast}$, Zeynep Akata$^{1,2}$\\
  $^{1}$University of Tübingen, $^{2}$MPI for Intelligent Systems\\
}

\begin{document}

\maketitle

\begin{abstract}
Large pre-trained, zero-shot capable models have shown considerable success both for standard transfer and adaptation tasks, with particular robustness towards distribution shifts.
In addition, subsequent fine-tuning can considerably improve performance on a selected downstream task. 
However, through naive fine-tuning, these zero-shot models lose their generalizability and robustness towards distribution shifts.
This is a particular problem for tasks such as Continual Learning (CL), where continuous adaptation has to be performed as new task distributions are introduced sequentially.
In this work, we showcase that where fine-tuning falls short to adapt such zero-shot capable models, simple momentum-based weight interpolation can provide consistent improvements for CL tasks in both memory-free and memory-based settings.
In particular, we find improvements of over $+4\%$ on standard CL benchmarks, while reducing the error to the upper limit of jointly training on all tasks at once in parts by more than half, allowing the continual learner to inch closer to the joint training limits.
\end{abstract}

\section{Introduction}
Continual Learning (CL) tackles the problem of learning from a non-stationary data stream, where training data is presented to the model not at once, but only in a sequence, and with limited capacity for retention and retraining. 
Not only does this require effective use of previously seen data, but also adaptation to novel context under continuously changing distribution shifts without catastrophic forgetting \cite{kirkpatrick_2017_ewc,rusu_2016_pnn,schwarz_2018_progresscompress,pmlr-v80-schwarz18a}.
Use cases are widespread, ranging from particularly compute-, time- or memory-limited to privacy-concerned applications \cite{lee2020clinical,buzzega2020dark,lesort2022long,wang_2022_prompt,ostapenko2022foundation}.

Consequentially, previous research has introduced a wide range of methods to address training under continual shifts, such as through the use of efficient data replay \cite{AGEM,buzzega2020dark,bang_2021_rainbow,prabhu_2020_gdumb}, regularization on the training dynamics \cite{kirkpatrick_2017_ewc,pmlr-v80-schwarz18a} or optimization procedures seeking for flat minima \cite{mirzadeh_2020_understanding,shi_2021_overcoming}.
Generally, these methods start from an untrained model which is then adapted to the data stream at hand.\\
While this has found practical success, more recently the use of large-scale pre-trained models ("foundation models" \cite{bommasani2021foundation,ostapenko2022foundation}) has become ubiquitous, as they have shown strong zero-shot generalizability to a variety of downstream tasks, with strong robustness to distribution shifts \cite{bommasani2021foundation}.

Their application to the CL problem set, which tackles a continuous distribution shift, stands to reason, with recent works showing notable benefits in the use of foundation models \cite{wang_2022_prompt,mehta2022an,ramasesh2022effect,wu2022pretrained}, particularly highlighting a reduction in catastrophic forgetting. 
Still, as learners are adapted to continuously shifting training distribution, even foundation models will suffer from forgetting through fine-tuning \cite{wortsman2022zeroshot}.

To maximize the benefits we can extract from the main continual learning process as well as the ability to classify novel samples at test time, it is thus important to minimize the impact on the generalizability of the adapted foundation model in order to account for potential further adaptations.

To allow for improved deployment in the CL setting, in this work we show how momentum-based weight-interpolation can help remedy issues of such models adapted in a continual fashion. 
In particular, as we want to maximally retain the generalizability of our adapted foundation model, we introduce a bifurcated adaptation mechanism by retaining an additional copy of the initial foundation model (denoted as \textit{slow} model). This slow model is excluded from the direct CL optimization process, and is only updated through linear momentum-interpolation with a task-adapted model copy (denoted as \textit{fast} model).

This is motivated by insights made in \cite{wortsman2022zeroshot}, who show that simple linear interpolation in weight space between the original zero-shot model and a variant fine-tuned to a task at hand allows for adaptation, while retaining better generalizability as compared to sole fine-tuning.
However, retaining a large collection of fine-tuned task expert models in the CL setting is memory intensive, impractical, and undesired. Instead, we show that we can simulate the empirical benefits highlighted in \cite{wortsman2022zeroshot} through repeated momentum interpolation between our foundation model and a continuously fine-tuned variant. This allows us to avoid the drawbacks of pure fine-tuning, while both specializing on the new stream of tasks, and retaining the generalizability of our foundation model.

Indeed, experiments on three standard CL benchmarks (Seq-CIFAR-10, Seq-CIFAR-100 and Seq-Tiny-ImageNet) show improvements in class- and task-incremental settings on both memory-based and memory-free methods by up to $+4\%$, and partly more than halving the error to the joint training performance bound. These results indicate that for practical usage of foundation models in a continuously distribution-shifting training scenario, momentum-based weight interpolation can be a reliable tool for consistent improvements that works well alongside any CL method.

\section{Related Work}
\textit{\textbf{Regularization-based methods}} augment the training objective to mitigate forgetting by keeping the current parameters close to previous task parameters, such as through moment matching \cite{rev_2} or Elastic Weight Consolidation (EWC) \cite{doi:10.1073/pnas.1611835114}, which performs Laplace approximations on the parameter posterior for each preceding task, using the means and covariances to regularize the current parameters via Mahalanobis distance minimization. Online Elastic Weight Consolidation (oEWC)\cite{pmlr-v80-schwarz18a} computes a momentum average of a single covariance matrix, and keeps the parameters from the last task only. Learning without Forgetting (LwF)\cite{8107520} also keeps the parameters from the last task and adds a cross-entropy term between logits computed with the old and current parameters, using data from the current task. \cite{mirzadeh_2020_dropout} show that dropout forces the model to learn a gating such that for different tasks, different paths of the network are active.\\

\textit{\textbf{Rehearsal-based methods}} utilize Experience Replay \cite{pmid2186426} \cite{doi:10.1080/09540099550039318} by storing a small subset of the training data into a buffer, and continually replaying it as the model moves on to learn new tasks. Dark Experience Replay (DER) \cite{NEURIPS2020_b704ea2c} introduces regularization in the rehearsal scheme by matching the logits of the past with the logits computed by the current network parameters. Gradient Episodic Memory (GEM) \cite{NIPS2017_f8752278} and Average Gradient Episodic Memory (A-GEM) \cite{chaudhry2018efficient} enforce optimization constraints in the current task using data from past tasks. GDumb \cite{prabhu2020greedy} greedily stores samples in memory, and only trains the model at test time using buffer data. DualNet \cite{NEURIPS2021_86a1fa88} uses a slow network for learning task-agnostic features through Self-Supervised Learning, and a fast network for learning task-specific features. Contrastive Continual Learning (Co2L) \cite{Cha_2021_ICCV} learns contrastive task-agnostic features, and trains a linear classifier using only buffer data. Our approach also bears conceptual similarities to the lookahead-style of optimization (see e.g. \cite{rev_1}), adapted to the continual learning problem.\\

\textit{\textbf{Flatness-seeking methods}} aim to operate in flat minima regions for each task in sequence, thereby retaining antecedent performance. Finding Flat Minima (F2M) \cite{NEURIPS2021_357cfba1} independently adds small random noise to the parameters, thereby obtaining similar but different loss functions which are optimized jointly in order to locate flat minima. \cite{NEURIPS2020_518a38cc} studies how batch size, dropout and learning rate decay affect the model's ability to find flat basins. \cite{mehta2021empirical} uses the Sharpness-Aware Minimization (SAM) \cite{foret2021sharpnessaware} procedure, which explicitly optimizes for parameters lying in flat basins. 

\section{Method}\label{sec:method}
In CL, a model $f_\theta$ is trained on a sequence of $T$ tasks, where for each task $t \in \{1, ..., T\}$ the learner only gets access to a subset of samples $D_t = \{(x_i, y_i)\}_{i=1}^{N_t}$, but is eventually evaluated on joint performance, i.e. we optimize

\begin{equation*}
\theta^{*} = \textstyle\argmin_\theta \textstyle\sum_{t=1}^T \mathbb{E}_{(x, y) \sim D_t} \left[L(f_\theta(x), y)\right].
\end{equation*}

The main challenge is that at task $t$, the model has no access to data from previous tasks $\tilde{t} \in \{1, ..., t-1\}$, therefore violating the typical IID data assumption. In this work, we investigate both the class-incremental setting, where subsets of classes are introduced in sequence, and the much easier task-incremental setting which jointly also provides respective task ids.

\subsection{Momentum-based Weight Interpolation for Continual Learning (MCL)}
To allow for effective and continuous adaptation of foundation models, we introduce momentum-based weight interpolation for CL. 
As our primary target is the retention of the generalizability and shift robustness of the underlying foundation model, it is important that minimal adaptation and fine-tuning is performed, while still allowing for a certain degree of adaptation to the target tasks at hand.
For that, we suggest a retention of a \textit{slow} model copy $\theta_\text{slow}$ which is kept disconnected from the entire adaptation process, while a second instantiation $\theta_\text{fast}$ is updated throughout the continual learning process.
As $\theta_\text{fast}$ adapts to the target distribution at hand, at every iteration we simultaneously perform an iterative updating on our slow weights through weight-space interpolation:

\begin{equation*}
    \theta_{\text{slow}} = \tau \cdot \theta_{\text{slow}} + (1-\tau) \cdot \theta_\text{fast} 
\end{equation*}

where $\tau$ is our \textit{momentum} hyperparameter. A simplified version of the procedure is summarized in Algorithm \ref{alg:mcl}. 
As this mechanism is task- and memory-agnostic with no dependence on task boundaries, it can be applied to any continual learning framework, both memory-based and memory-free.
And while straightforward and simple, the intuitively better retention of foundation model weights in the continual learning setting is well motivated.

Beyond a conceptual connection to the Complimentary Learning Systems (CLS) \cite{pmid7624455,Cha_2021_co2l} theory from neuroscience which depicts human continual learning as an interplay of a fast adaptive and a slow retentive system, on a methodological level  \cite{izmailov2018averaging} show that maintaining a running average of weights leads to wider optima and retained generalization during the standard fine-tuning process of a pre-trained model. 

In addition, \cite{wortsman2022zeroshot} showcase that zero-shot and fine-tuned model weights are often connected by a linear path which retains performance. 
It therefore stands to reason that our linear momentum-based interpolation across task iterations allows us to connect to the performance of our task-adapted fast variant, while maintaining the generalizability our foundation model weights $\theta_\text{slow}$. The consequently sustained implicit optimization for a flatter minimum around $\theta_\text{slow}$, which is only updated through momentum-based interpolation, has strong ties to improved generalization across task sequences in continual learning \cite{NEURIPS2021_357cfba1,NEURIPS2020_518a38cc,foret2021sharpnessaware}, which we see reflected in our benchmark experiments in the next section.

\begin{algorithm}
\caption{Momentum-based Weight Interpolation for Continual Learning (MCL)}\label{alg:mcl}
\begin{algorithmic}[1]
\Require Pre-trained weights $\theta_{pre}$, Momentum $\tau \in [0, 1]$ 
\State $\theta_\text{fast} \gets \theta_{pre}$
\State $\theta_{\text{slow}} \gets \theta_{pre}$
\For{$t \gets 1$ \ldots num\_tasks}
    \For{$e \gets 1$ \ldots num\_epochs}   
        \For{$(x, y) \sim D_t$}   
            \State{$\theta_\text{fast} \gets \theta_\text{fast} - \alpha \nabla \mathcal{L}(f_{\theta_\text{fast}}(x), y)$}
            \State{$\theta_{\text{slow}} \gets \tau \cdot \theta_{\text{slow}} + (1-\tau) \cdot \theta_\text{fast}$}
        \EndFor
    \EndFor
\EndFor
\State $\theta_\text{fast} \gets \theta_{slow}$
\end{algorithmic}
\end{algorithm}

\section{Experiments}
\label{sec:experiments}

\textbf{Datasets.} We evaluate our method on three datasets commonly used in the literature: CIFAR-10 \cite{Krizhevsky09learningmultiple}, CIFAR-100 \cite{Krizhevsky09learningmultiple}, and Tiny ImageNet. We split each dataset into several tasks of non-overlapping classes: Seq-CIFAR-10 consisting of 5 tasks (2 classes each) and Seq-CIFAR-100/Seq-Tiny-ImageNet consisting of 10 tasks (10 and 20 classes each, respectively).\\ 

\textbf{Training.} For our zero-shot model we use a pre-trained CLIP ViT-B/16 \cite{pmlr-v139-radford21a}. We built our CL experiments on \cite{NEURIPS2020_b704ea2c} which implements several CL benchmarks in PyTorch \cite{NEURIPS2019_9015}. All methods follow a standardized training protocol - trained on Nvidia 2080Ti's using SGD \cite{pmlr-v28-shamir13}, a fixed learning rate and no scheduler, with the same fine-tuning budget of $10$ epochs. We perform grid searches on a random train subset to select the best learning rate $\alpha\in\{10^{-2}, 10^{-3}, 10^{-4}, 10^{-5}, 10^{-6}, 10^{-7}\}$ as well as the best momentum strength $\tau\in\{0.995, 0.997, 0.999, 0.9995, 0.9997, 0.9999 \}$. We refer the reader to the appendix (\S\ref{sec:ablation-study}) for an ablation study of the hyperparameters.\\

\textbf{Evaluation.}  For both Task Incremental Learning (Task-IL) and Class Incremental Learning (Class-IL) scenarios, we report the final classification accuracy over all encountered classes, with task identities also provided in the Task-IL setting (making it a noticeably easier problem to solve). 

\begin{table*}[h!]
  \caption{Baselines}
  \label{table-baselines}
  \centering
\centering  
  \begin{tabular}{cccc}
    \toprule
    \textbf{Baseline} & \textbf{CIFAR-10} & \textbf{CIFAR-100} & \textbf{Tiny-ImageNet} \\ 
    \midrule 
    ZERO-SHOT & $88.77$ & $63.11$ & $58.53$ \\ 
    JOINT & $97.53 \pm 0.08$ & $87.22 \pm 0.54$ & $78.86 \pm 1.38$ \\
    \bottomrule
  \end{tabular}
\end{table*}

\subsection{Experimental Results}  \label{sec:experimental-results}
In this section, we experiment with the use of momentum-based weight interpolation in three standard CL method categories: fine-tuning (pure SGD \cite{pmlr-v28-shamir13}), regularization-based (oEWC \cite{pmlr-v80-schwarz18a}), and rehearsal-based (DER++ \cite{NEURIPS2020_b704ea2c} with buffer size $500$ and $5000$). 

The results presented below are obtained over three seeds, alongside which we provide the zero-shot lower bound (Tab. \ref{table-baselines}).
Interestingly, the non-adapted zero-shot performance already in parts vastly outperforms comparable adaptation with state-of-the-art methods not relying on foundation models, with e.g. DER++ \cite{NEURIPS2020_b704ea2c} reporting $72.70\pm1.36\%$ with a buffer of 500, and $85.40\pm0.49\%$ with a buffer of 5000 on CIFAR-10, while zero-shot performance of our foundation model already achieves $88.77\%$. This difference is even further exacerbated on Tiny-ImageNet, with $19.38\pm1.41\%$ and $39.02\pm0.97$ for buffer sizes of 500 and 5000 respectively, versus 58.53\% for zero-shot performance, verifying the potential \cite{mehta2022an,ramasesh2022effect} of foundation models in CL.

To provide an upper bound, we train on all tasks jointly (Tab. \ref{table-baselines}). 
Since joint training is evaluated without task boundaries, this upper bound does not hold for Task-IL scenarios.
Next, in Tab. \ref{table-continual} we present the results on the CL benchmarks. We empirically show that, as motivated in Sec. \ref{sec:method}, keeping a momentum-interpolated version of the foundation model results in consistent improvements.

In particular, our results show that adaptation to the task distribution at hand is beneficial even with simple fine-tuning. Even when accounting for a change in learning rate (as noted in \S\ref{sec:experiments} and done for every baseline), we find that additional momentum-based weight interpolation offers consistent benefits in both class- and task-incremental settings, with nearly $+4\%$ improvement on both Seq-CIFAR-100 and Seq-Tiny-ImageNet. Furthermore, through momentum-updating, we can push simple fine-tuning close or even over the performance of a state-of-the-art CL framework (DER++).
Additionally, we observe similar performance improvements even when applied on top of separate CL frameworks, both memory-free (oEWC, e.g. $74.07\pm0.20\rightarrow 77.25\pm0.31$ on Seq-CIFAR-100) and memory-based (DER++ with 500 memory samples, $76.78\pm0.23\rightarrow 82.01\pm0.31$).

Interestingly, a momentum-extended DER++ with a buffer size of 500 also almost closes the gap in performance to the non-momentum based DER++ with a much larger buffer size 5000, which, even with such a large memory, also sees significant improvements on the particularly more complex CL tasks (Seq-Tiny-ImageNet, $76.54\pm0.10\rightarrow78.26\pm0.14$).

This demonstrates that the need for buffer sizes in CL frameworks built around foundation models can decrease significantly (in this case, 10-fold) through momentum-based weight-space interpolation.
We do note that while not necessary for the benchmarks at hand, longer task sequence may benefit from a re-synchronization of $\theta_\text{slow}$ and $\theta_\text{fast}$.

Finally, we find that momentum-based DER++ with a buffer of 5000 even further closes the gap to the joint optimization upper bound - looking at the error, we find a drop of $0.45\%\rightarrow 0.32\%$ on Seq-CIFAR-10, $4.06\%\rightarrow2.28\%$ on Seq-CIFAR-100, and $2.32\%\rightarrow0.6\%$ on Seq-Tiny-ImageNet, which marks a nearly $75\%$ reduction. 
Conclusively, these results indicate the significant benefits of retaining a momentum-updated model copy when introducing foundation models into the CL setting, both for consistent relative improvements, but also to minimize the performance drop when moving from the standard joint optimization to a continual learning scenario.

\begin{table*}[t!]
  \caption{Continual Learning setting -- training and evaluating on sequences of tasks.}
  \label{table-continual}
  \centering
\centering\resizebox{1\textwidth}{!}{  
  \begin{tabular}{cccccccccc}
    \toprule
    \multirow{2}{*}{\textbf{Method}} & \multirow{2}{*}{\textbf{Momentum}}  & \multicolumn{2}{c}{\textbf{Seq-CIFAR-10}}  & \multicolumn{2}{c}{\textbf{Seq-CIFAR-100}}  & \multicolumn{2}{c}{\textbf{Seq-Tiny-ImageNet}} \\
    & & Class-IL & Task-IL & Class-IL & Task-IL & Class-IL & Task-IL \\
    \midrule 
    
    \multirow{2}{*}{SGD} & no & $91.38 \pm 0.04$ & $98.17 \pm 0.01$ & $74.36 \pm 0.03$ & $93.59 \pm 0.04$ & $67.30 \pm 0.08$ & $82.12 \pm 0.07$ \\
     & yes & $92.46 \pm 0.11$ & $98.43 \pm 0.01$ & $77.52 \pm 0.37$ & $94.98 \pm 0.17$ & $71.09 \pm 0.28$ & $85.22 \pm 0.32$ \\
    \midrule

    \multirow{2}{*}{oEWC} & no & $90.67 \pm 0.01$ & $98.17 \pm 0.01$ & $74.07 \pm 0.20$ & $93.80 \pm 0.02$ & $66.60 \pm 0.02$ & $81.79 \pm 0.02$ \\
     & yes & $91.87 \pm 0.57$ & $98.88 \pm 0.12$ & $77.25 \pm 0.31$ & $95.09 \pm 0.01$ & $71.57 \pm 0.05$ & $85.94 \pm 0.07$ \\
    \midrule

    \multirow{2}{*}{DER++ (500)} & no & $94.65 \pm 0.16$ & $99.38 \pm 0.10$ & $76.68 \pm 0.23$ & $95.05 \pm 0.09$ & $71.05 \pm 0.12$ & $84.42 \pm 0.22$ \\
     & yes & $95.73 \pm 0.21$ & $99.50 \pm 0.04$ & $82.01 \pm 0.31$ & $96.69 \pm 0.03$ & $75.11 \pm 0.02$ & $87.80 \pm 0.27$ \\
    \midrule

    \multirow{2}{*}{DER++ (5000)} & no & $97.08 \pm 0.04$ & $99.60 \pm 0.01$ & $83.16 \pm 0.20$ & $97.03 \pm 0.11$ & $76.54 \pm 0.10$ & $88.44 \pm 0.04$ \\
     & yes & $97.21 \pm 0.11$ & $99.62 \pm 0.01$ & $84.94 \pm 0.07$ & $97.13 \pm 0.05$ & $78.26 \pm 0.14$ & $89.00 \pm 0.11$ \\

    \bottomrule
  \end{tabular}}
\end{table*}

\section{Conclusion}
This work tackles the adaptation of large-scale pre-trained zero-shot models to continual learning (CL). To retain the strong generalizability and robustness of these models even under continuous fine-tuning, we propose the use of a momentum-based interpolation between a slow-moving zero-shot model excluded from the direct CL process and a task-adapted fast variant. Through this simple extension, we find consistent improvements in performance across three standard CL benchmarks (Seq-CIFAR-10, Seq-CIFAR-100, Seq-Tiny-ImageNet) on both memory-based and memory-free approaches, of in parts more than $+4\%$. In addition, we find the distance between continual learning and joint task optimization performance in some cases to even be more than halved. Based on these insights, the generalizability of large-scale pre-trained zero-shot models, and the simplicity of the proposed setup, we believe the adoption of our approach to be of high practical interest.

\section*{Acknowledgements}
Karsten Roth thanks the International Max Planck Research School as well as the European Laboratory for Learning and Intelligent Systems (ELLIS) PhD program for support. Zeynep Akata acknowledges partial funding by the ERC (853489 - DEXIM) and DFG (2065/1 - Project number 390727645) under Germany's Excellence Strategy.

\small
\bibliographystyle{ieee_fullname}
\bibliography{main}

\newpage
\appendix

\section{Appendix}

\begin{figure*}[ht!]
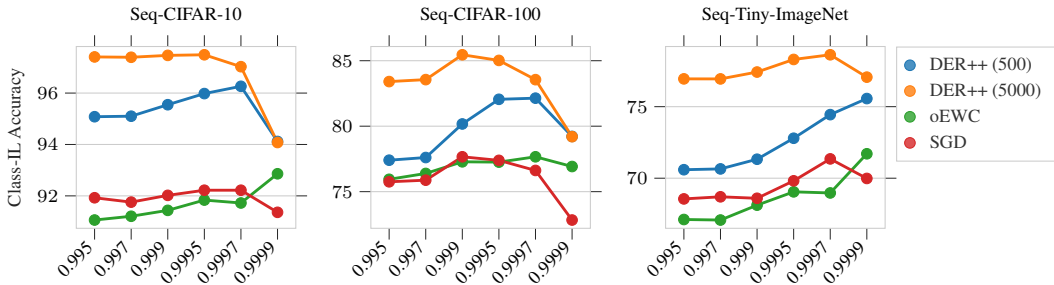

    \centering
    \include{figures/tau_class-il-acc}
    \vspace{-1cm}
    \caption{The effect of the hyper-parameter $\tau$ on the Class-IL Accuracy}%
    \label{fig:tau_class-il-acc}%
\end{figure*}

\subsection{Ablation study}
\label{sec:ablation-study}
\textbf{Momentum strength. } In Figure \ref{fig:tau_class-il-acc} we show how the momentum strength $\tau$ affects the model's Class-IL Accuracy. While we find that the optimal value of $\tau$ is dataset-dependent, it is encouraging that the vastly different methods show surprisingly similar behavior for a given dataset.

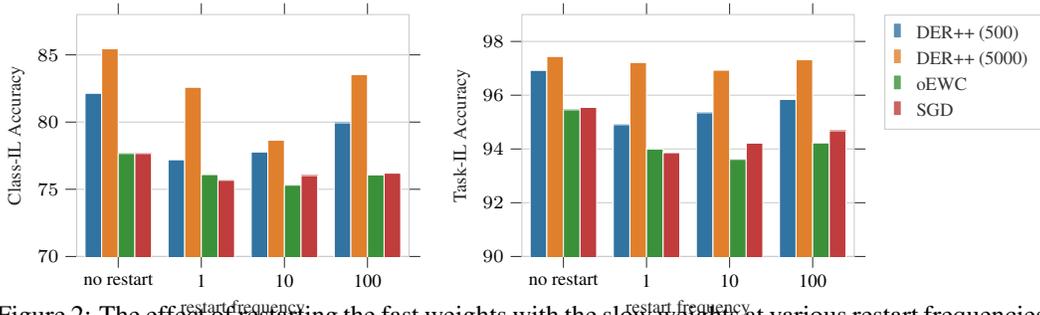
\begin{figure*}[ht!]
    \centering
\begin{tikzpicture}
\scriptsize

\definecolor{brown1926061}{RGB}{192,60,61}
\definecolor{darkslategray38}{RGB}{38,38,38}
\definecolor{darkslategray66}{RGB}{66,66,66}
\definecolor{lightgray204}{RGB}{204,204,204}
\definecolor{peru22412844}{RGB}{224,128,44}
\definecolor{seagreen5814558}{RGB}{58,145,58}
\definecolor{steelblue49115161}{RGB}{49,115,161}

\begin{groupplot}[
    group style={group size=2 by 1},
    width=6cm,
    height=4.8cm
]

\nextgroupplot[
axis line style={lightgray204},
legend cell align={left},
legend style={
  fill opacity=0.8,
  draw opacity=1,
  text opacity=1,
  at={(2.43,1)},
  anchor=north west,
  draw=lightgray204
},
tick align=outside,
unbounded coords=jump,
x grid style={lightgray204},
xlabel=\textcolor{darkslategray38}{restart frequency},
xmajorticks=true,
xmin=-0.5, xmax=3.5,
xtick style={color=darkslategray38},
xtick={0,1,2,3},
xticklabels={\text{no restart},1,10,100},
y grid style={lightgray204},
ylabel=\textcolor{darkslategray38}{Class-IL Accuracy},
ymajorgrids,
ymajorticks=true,
ymin=70, ymax=88,
ytick style={color=darkslategray38}
]
\draw[draw=white,fill=steelblue49115161] (axis cs:-0.4,0) rectangle (axis cs:-0.2,82.14);
\addlegendimage{ybar,ybar legend,draw=white,fill=steelblue49115161}
\addlegendentry{DER++ (500)}

\draw[draw=white,fill=steelblue49115161] (axis cs:0.6,0) rectangle (axis cs:0.8,77.1933333333333);
\draw[draw=white,fill=steelblue49115161] (axis cs:1.6,0) rectangle (axis cs:1.8,77.77);
\draw[draw=white,fill=steelblue49115161] (axis cs:2.6,0) rectangle (axis cs:2.8,79.9433333333333);
\draw[draw=white,fill=peru22412844] (axis cs:-0.2,0) rectangle (axis cs:0,85.4533333333333);
\addlegendimage{ybar,ybar legend,draw=white,fill=peru22412844}
\addlegendentry{DER++ (5000)}

\draw[draw=white,fill=peru22412844] (axis cs:0.8,0) rectangle (axis cs:1,82.5866666666667);
\draw[draw=white,fill=peru22412844] (axis cs:1.8,0) rectangle (axis cs:2,78.6533333333333);
\draw[draw=white,fill=peru22412844] (axis cs:2.8,0) rectangle (axis cs:3,83.53);
\draw[draw=white,fill=seagreen5814558] (axis cs:2.77555756156289e-17,0) rectangle (axis cs:0.2,77.6633333333333);
\addlegendimage{ybar,ybar legend,draw=white,fill=seagreen5814558}
\addlegendentry{oEWC}

\draw[draw=white,fill=seagreen5814558] (axis cs:1,0) rectangle (axis cs:1.2,76.0933333333333);
\draw[draw=white,fill=seagreen5814558] (axis cs:2,0) rectangle (axis cs:2.2,75.3066666666667);
\draw[draw=white,fill=seagreen5814558] (axis cs:3,0) rectangle (axis cs:3.2,76.07);
\draw[draw=white,fill=brown1926061] (axis cs:0.2,0) rectangle (axis cs:0.4,77.6633333333333);
\addlegendimage{ybar,ybar legend,draw=white,fill=brown1926061}
\addlegendentry{SGD}

\draw[draw=white,fill=brown1926061] (axis cs:1.2,0) rectangle (axis cs:1.4,75.67);
\draw[draw=white,fill=brown1926061] (axis cs:2.2,0) rectangle (axis cs:2.4,76.0233333333333);
\draw[draw=white,fill=brown1926061] (axis cs:3.2,0) rectangle (axis cs:3.4,76.2);
\addplot [line width=1.08pt, darkslategray66]
table {%
-0.3 nan
-0.3 nan
};
\addplot [line width=1.08pt, darkslategray66]
table {%
0.7 nan
0.7 nan
};
\addplot [line width=1.08pt, darkslategray66]
table {%
1.7 nan
1.7 nan
};
\addplot [line width=1.08pt, darkslategray66]
table {%
2.7 nan
2.7 nan
};
\addplot [line width=1.08pt, darkslategray66]
table {%
-0.1 nan
-0.1 nan
};
\addplot [line width=1.08pt, darkslategray66]
table {%
0.9 nan
0.9 nan
};
\addplot [line width=1.08pt, darkslategray66]
table {%
1.9 nan
1.9 nan
};
\addplot [line width=1.08pt, darkslategray66]
table {%
2.9 nan
2.9 nan
};
\addplot [line width=1.08pt, darkslategray66]
table {%
0.1 nan
0.1 nan
};
\addplot [line width=1.08pt, darkslategray66]
table {%
1.1 nan
1.1 nan
};
\addplot [line width=1.08pt, darkslategray66]
table {%
2.1 nan
2.1 nan
};
\addplot [line width=1.08pt, darkslategray66]
table {%
3.1 nan
3.1 nan
};
\addplot [line width=1.08pt, darkslategray66]
table {%
0.3 nan
0.3 nan
};
\addplot [line width=1.08pt, darkslategray66]
table {%
1.3 nan
1.3 nan
};
\addplot [line width=1.08pt, darkslategray66]
table {%
2.3 nan
2.3 nan
};
\addplot [line width=1.08pt, darkslategray66]
table {%
3.3 nan
3.3 nan
};

\nextgroupplot[
axis line style={lightgray204},
xshift=0.5cm,
tick align=outside,
unbounded coords=jump,
x grid style={lightgray204},
xlabel=\textcolor{darkslategray38}{restart frequency},
xmajorticks=true,
xmin=-0.5, xmax=3.5,
xtick style={color=darkslategray38},
xtick={0,1,2,3},
xticklabels={\text{no restart},1,10,100},
y grid style={lightgray204},
ylabel=\textcolor{darkslategray38}{Task-IL Accuracy},
ymajorgrids,
ymajorticks=true,
ymin=90, ymax=99,
ytick style={color=darkslategray38}
]
\draw[draw=white,fill=steelblue49115161] (axis cs:-0.4,0) rectangle (axis cs:-0.2,96.9233333333333);
\draw[draw=white,fill=steelblue49115161] (axis cs:0.6,0) rectangle (axis cs:0.8,94.9033333333333);
\draw[draw=white,fill=steelblue49115161] (axis cs:1.6,0) rectangle (axis cs:1.8,95.3466666666667);
\draw[draw=white,fill=steelblue49115161] (axis cs:2.6,0) rectangle (axis cs:2.8,95.8433333333333);
\draw[draw=white,fill=peru22412844] (axis cs:-0.2,0) rectangle (axis cs:0,97.44);
\draw[draw=white,fill=peru22412844] (axis cs:0.8,0) rectangle (axis cs:1,97.2166666666667);
\draw[draw=white,fill=peru22412844] (axis cs:1.8,0) rectangle (axis cs:2,96.93);
\draw[draw=white,fill=peru22412844] (axis cs:2.8,0) rectangle (axis cs:3,97.32);
\draw[draw=white,fill=seagreen5814558] (axis cs:2.77555756156289e-17,0) rectangle (axis cs:0.2,95.4566666666667);
\draw[draw=white,fill=seagreen5814558] (axis cs:1,0) rectangle (axis cs:1.2,93.9966666666667);
\draw[draw=white,fill=seagreen5814558] (axis cs:2,0) rectangle (axis cs:2.2,93.6133333333333);
\draw[draw=white,fill=seagreen5814558] (axis cs:3,0) rectangle (axis cs:3.2,94.2233333333333);
\draw[draw=white,fill=brown1926061] (axis cs:0.2,0) rectangle (axis cs:0.4,95.5466666666667);
\draw[draw=white,fill=brown1926061] (axis cs:1.2,0) rectangle (axis cs:1.4,93.86);
\draw[draw=white,fill=brown1926061] (axis cs:2.2,0) rectangle (axis cs:2.4,94.2166666666667);
\draw[draw=white,fill=brown1926061] (axis cs:3.2,0) rectangle (axis cs:3.4,94.6766666666667);
\addplot [line width=1.08pt, darkslategray66, forget plot]
table {%
-0.3 nan
-0.3 nan
};
\addplot [line width=1.08pt, darkslategray66, forget plot]
table {%
0.7 nan
0.7 nan
};
\addplot [line width=1.08pt, darkslategray66, forget plot]
table {%
1.7 nan
1.7 nan
};
\addplot [line width=1.08pt, darkslategray66, forget plot]
table {%
2.7 nan
2.7 nan
};
\addplot [line width=1.08pt, darkslategray66, forget plot]
table {%
-0.1 nan
-0.1 nan
};
\addplot [line width=1.08pt, darkslategray66, forget plot]
table {%
0.9 nan
0.9 nan
};
\addplot [line width=1.08pt, darkslategray66, forget plot]
table {%
1.9 nan
1.9 nan
};
\addplot [line width=1.08pt, darkslategray66, forget plot]
table {%
2.9 nan
2.9 nan
};
\addplot [line width=1.08pt, darkslategray66, forget plot]
table {%
0.1 nan
0.1 nan
};
\addplot [line width=1.08pt, darkslategray66, forget plot]
table {%
1.1 nan
1.1 nan
};
\addplot [line width=1.08pt, darkslategray66, forget plot]
table {%
2.1 nan
2.1 nan
};
\addplot [line width=1.08pt, darkslategray66, forget plot]
table {%
3.1 nan
3.1 nan
};
\addplot [line width=1.08pt, darkslategray66, forget plot]
table {%
0.3 nan
0.3 nan
};
\addplot [line width=1.08pt, darkslategray66, forget plot]
table {%
1.3 nan
1.3 nan
};
\addplot [line width=1.08pt, darkslategray66, forget plot]
table {%
2.3 nan
2.3 nan
};
\addplot [line width=1.08pt, darkslategray66, forget plot]
table {%
3.3 nan
3.3 nan
};
\end{groupplot}

\end{tikzpicture}
    \vspace{-1cm}
    \caption{The effect of restarting the fast weights with the slow weights at various restart frequencies. }%
    \label{fig:restats}%
\end{figure*}

\textbf{Restart frequency. } Next, we examine whether it is beneficial to restart the fast weights $\theta_\text{fast}$ with the slow weights $\theta_{\text{slow}}$ during training (instead of only at the end as per default, i.e. Line 11 in Algorithm \ref{alg:mcl}). To this end, we introduce a new hyperparameter \textit{restart frequency} which specifies after how many gradient steps we perform a restart. From the results detailed in Figure \ref{fig:restats}, we find that restarting the fast weights is not beneficial to the generalization performance.

\textbf{Update frequency. } Finally, we examine whether it is beneficial to perform the update of the slow weights (Line 7 in Algorithm \ref{alg:mcl}) at various frequencies. For this purpose, we introduce a new hyperparameter \textit{update frequency} which specifies after how many gradient steps we update the slow weights. From the results summarized in Figure \ref{fig:updates}, we find that updating at frequencies higher than 1 (where 1 is the default behavior of our algorithm) does not provide a boost in performance.



\begin{figure*}[hb!]
    \centering
\begin{tikzpicture}
\scriptsize

\definecolor{brown1926061}{RGB}{192,60,61}
\definecolor{darkslategray38}{RGB}{38,38,38}
\definecolor{darkslategray66}{RGB}{66,66,66}
\definecolor{lightgray204}{RGB}{204,204,204}
\definecolor{peru22412844}{RGB}{224,128,44}
\definecolor{seagreen5814558}{RGB}{58,145,58}
\definecolor{steelblue49115161}{RGB}{49,115,161}

\begin{groupplot}[
    group style={group size=2 by 1},
    width=6cm,
    height=4.8cm
]

\nextgroupplot[
axis line style={lightgray204},
legend cell align={left},
legend style={
  fill opacity=0.8,
  draw opacity=1,
  text opacity=1,
  at={(2.43,1)},
  anchor=north west,
  draw=lightgray204
},
tick align=outside,
unbounded coords=jump,
x grid style={lightgray204},
xlabel=\textcolor{darkslategray38}{update frequency},
xmajorticks=true,
xmin=-0.5, xmax=2.5,
xtick style={color=darkslategray38},
xtick={0,1,2},
xticklabels={1,10,100},
y grid style={lightgray204},
ylabel=\textcolor{darkslategray38}{Class-IL Accuracy},
ymajorgrids,
ymajorticks=true,
ymin=70, ymax=88,
ytick style={color=darkslategray38}
]
\draw[draw=white,fill=steelblue49115161] (axis cs:-0.4,0) rectangle (axis cs:-0.2,82.14);
\addlegendimage{ybar,ybar legend,draw=white,fill=steelblue49115161}
\addlegendentry{DER++ (500)}

\draw[draw=white,fill=steelblue49115161] (axis cs:0.6,0) rectangle (axis cs:0.8,82.1766666666667);
\draw[draw=white,fill=steelblue49115161] (axis cs:1.6,0) rectangle (axis cs:1.8,76.9233333333333);
\draw[draw=white,fill=peru22412844] (axis cs:-0.2,0) rectangle (axis cs:0,85.4533333333333);
\addlegendimage{ybar,ybar legend,draw=white,fill=peru22412844}
\addlegendentry{DER++ (5000)}

\draw[draw=white,fill=peru22412844] (axis cs:0.8,0) rectangle (axis cs:1,85.0466666666667);
\draw[draw=white,fill=peru22412844] (axis cs:1.8,0) rectangle (axis cs:2,77.2666666666667);
\draw[draw=white,fill=seagreen5814558] (axis cs:2.77555756156289e-17,0) rectangle (axis cs:0.2,77.6633333333333);
\addlegendimage{ybar,ybar legend,draw=white,fill=seagreen5814558}
\addlegendentry{oEWC}

\draw[draw=white,fill=seagreen5814558] (axis cs:1,0) rectangle (axis cs:1.2,77.7066666666667);
\draw[draw=white,fill=seagreen5814558] (axis cs:2,0) rectangle (axis cs:2.2,75.0733333333333);
\draw[draw=white,fill=brown1926061] (axis cs:0.2,0) rectangle (axis cs:0.4,77.6633333333333);
\addlegendimage{ybar,ybar legend,draw=white,fill=brown1926061}
\addlegendentry{SGD}

\draw[draw=white,fill=brown1926061] (axis cs:1.2,0) rectangle (axis cs:1.4,77.3833333333333);
\draw[draw=white,fill=brown1926061] (axis cs:2.2,0) rectangle (axis cs:2.4,71.75);
\addplot [line width=1.08pt, darkslategray66]
table {%
-0.3 nan
-0.3 nan
};
\addplot [line width=1.08pt, darkslategray66]
table {%
0.7 nan
0.7 nan
};
\addplot [line width=1.08pt, darkslategray66]
table {%
1.7 nan
1.7 nan
};
\addplot [line width=1.08pt, darkslategray66]
table {%
-0.1 nan
-0.1 nan
};
\addplot [line width=1.08pt, darkslategray66]
table {%
0.9 nan
0.9 nan
};
\addplot [line width=1.08pt, darkslategray66]
table {%
1.9 nan
1.9 nan
};
\addplot [line width=1.08pt, darkslategray66]
table {%
0.1 nan
0.1 nan
};
\addplot [line width=1.08pt, darkslategray66]
table {%
1.1 nan
1.1 nan
};
\addplot [line width=1.08pt, darkslategray66]
table {%
2.1 nan
2.1 nan
};
\addplot [line width=1.08pt, darkslategray66]
table {%
0.3 nan
0.3 nan
};
\addplot [line width=1.08pt, darkslategray66]
table {%
1.3 nan
1.3 nan
};
\addplot [line width=1.08pt, darkslategray66]
table {%
2.3 nan
2.3 nan
};

\nextgroupplot[
axis line style={lightgray204},
xshift=0.5cm,
tick align=outside,
unbounded coords=jump,
x grid style={lightgray204},
xlabel=\textcolor{darkslategray38}{update frequency},
xmajorticks=true,
xmin=-0.5, xmax=2.5,
xtick style={color=darkslategray38},
xtick={0,1,2},
xticklabels={1,10,100},
y grid style={lightgray204},
ylabel=\textcolor{darkslategray38}{Task-IL Accuracy},
ymajorgrids,
ymajorticks=true,
ymin=90, ymax=99,
ytick style={color=darkslategray38}
]
\draw[draw=white,fill=steelblue49115161] (axis cs:-0.4,0) rectangle (axis cs:-0.2,96.9233333333333);
\draw[draw=white,fill=steelblue49115161] (axis cs:0.6,0) rectangle (axis cs:0.8,96.9166666666667);
\draw[draw=white,fill=steelblue49115161] (axis cs:1.6,0) rectangle (axis cs:1.8,94.86);
\draw[draw=white,fill=peru22412844] (axis cs:-0.2,0) rectangle (axis cs:0,97.44);
\draw[draw=white,fill=peru22412844] (axis cs:0.8,0) rectangle (axis cs:1,97.4733333333333);
\draw[draw=white,fill=peru22412844] (axis cs:1.8,0) rectangle (axis cs:2,95.27);
\draw[draw=white,fill=seagreen5814558] (axis cs:2.77555756156289e-17,0) rectangle (axis cs:0.2,95.4566666666667);
\draw[draw=white,fill=seagreen5814558] (axis cs:1,0) rectangle (axis cs:1.2,95.4566666666667);
\draw[draw=white,fill=seagreen5814558] (axis cs:2,0) rectangle (axis cs:2.2,93.8933333333333);
\draw[draw=white,fill=brown1926061] (axis cs:0.2,0) rectangle (axis cs:0.4,95.5466666666667);
\draw[draw=white,fill=brown1926061] (axis cs:1.2,0) rectangle (axis cs:1.4,95.4033333333334);
\draw[draw=white,fill=brown1926061] (axis cs:2.2,0) rectangle (axis cs:2.4,91.72);
\addplot [line width=1.08pt, darkslategray66, forget plot]
table {%
-0.3 nan
-0.3 nan
};
\addplot [line width=1.08pt, darkslategray66, forget plot]
table {%
0.7 nan
0.7 nan
};
\addplot [line width=1.08pt, darkslategray66, forget plot]
table {%
1.7 nan
1.7 nan
};
\addplot [line width=1.08pt, darkslategray66, forget plot]
table {%
-0.1 nan
-0.1 nan
};
\addplot [line width=1.08pt, darkslategray66, forget plot]
table {%
0.9 nan
0.9 nan
};
\addplot [line width=1.08pt, darkslategray66, forget plot]
table {%
1.9 nan
1.9 nan
};
\addplot [line width=1.08pt, darkslategray66, forget plot]
table {%
0.1 nan
0.1 nan
};
\addplot [line width=1.08pt, darkslategray66, forget plot]
table {%
1.1 nan
1.1 nan
};
\addplot [line width=1.08pt, darkslategray66, forget plot]
table {%
2.1 nan
2.1 nan
};
\addplot [line width=1.08pt, darkslategray66, forget plot]
table {%
0.3 nan
0.3 nan
};
\addplot [line width=1.08pt, darkslategray66, forget plot]
table {%
1.3 nan
1.3 nan
};
\addplot [line width=1.08pt, darkslategray66, forget plot]
table {%
2.3 nan
2.3 nan
};
\end{groupplot}

\end{tikzpicture}
    \vspace{-1cm}
    \caption{The effect of computing the momentum update at various update  frequencies.}%
    \label{fig:updates}%
\end{figure*}


\end{document}